# Velocity/Position Integration Formula (II): Application to Strapdown Inertial Navigation Computation

Yuanxin Wu and Xianfei Pan

**Abstract— Inertial navigation applications are usually referenced to a rotating frame. Consideration of the navigation reference frame rotation in the inertial navigation algorithm design is an important but so far less seriously treated issue, especially for the future ultra-precision navigation system of several meters per hour. This paper proposes a rigorous approach to tackle the issue of navigation frame rotation in velocity/position computation by use of the newly-devised velocity/position integration formulae in the Part I companion paper. The two integration formulae set a well-founded cornerstone for the velocity/position algorithms design that makes the comprehension of the inertial navigation computation principle more accessible to practitioners, and different approximations to the integrals involved will give birth to various velocity/position update algorithms. Two-sample velocity and position algorithms are derived to exemplify the design process. In the context of level-flight airplane examples, the derived algorithm is analytically and numerically compared to the typical algorithms existing in the literature. The results throw light on the problems in existing algorithms and the potential benefits of the derived algorithm.**

## I. Introduction

Over fifty years have seen the fruitful algorithm development for the strapdown inertial navigation system using a triad of gyroscopes and accelerometers that are strapped down to the vehicle [1]. Because the inertial sensors rotate

This work was supported in part by the Fok Ying Tung Foundation (131061), National Natural Science Foundation of China (61174002), the Foundation for the Author of National Excellent Doctoral Dissertation of People's Republic of China (FANEDD 200897) and Program for New Century Excellent Talents in University (NCET-10-0900).

Authors' address: Department of Automatic Control, College of Mechatronics and Automation, National University of Defense Technology, Changsha, Hunan, P. R. China, 410073. Tel/Fax: 086-0731-84576305-8212 (e-mail: yuanx_wu@hotmail.com).

along with the vehicle, the strapdown navigation algorithm is much more complex than that for gimbaled inertial navigation, even worse when a rotating navigation frame is chosen [2-5]. The algorithm researches have been mostly focused on body-related dynamic attitude/velocity/position computation, leading to the state-of-the-art coning/sculling/scrolling corrections that form the basis of the modern strapdown inertial navigation algorithm [1, 4-9]. Though widely believed that the strapdown algorithm has been currently more than adequate, accuracy pursuit is well motivated and objectively necessitated for the on-the-horizon ultra-precision inertial navigation system of several meters per hour [10].

Navigation applications are usually referenced to a rotating frame, e.g., the Earth frame or the local level frame. Consideration of the navigation frame rotation in the velocity/position computation is an important algorithm issue for the future ultra-precision navigation system [10], which has so far been less seriously handled in the literature and text books [1-3, 5]. Regarding the attitude computation, the issue can be addressed by the combination of the body frame and the rotating navigation frame both relative to some chosen inertial frame [4]. When it comes to the velocity/position computation, the issue is not as simple as most thought, due to the single/double integrations of the transformed specific force in the velocity/position calculation. To our best knowledge, the first trial treatment was by Savage in [11] with yet little details and then the issue was further investigated in [1, 5] under coarse assumptions.

This paper is aimed to propose a rigorous approach to tackle the issue of the navigation frame rotation for velocity and position computation. It is achieved by use of the velocity integration formula and the position integration formula that are newly derived and applied to the in-flight alignment in the companion paper [12]. Hopefully, the paper will benefit the comprehension of the inertial navigation computation principle[1] and provide a well-founded systematic framework to design the velocity and position updating algorithms with potentially improved accuracy. This paper, along with the companion paper [12], connects the navigation computation problem and the alignment problem in an interesting way.

The remaining of the paper is organized as follows. Section II revisits the velocity/position integration formulae in the incremental form, which sets a solid basis for the velocity/position algorithms design. Section III demonstrates the two-sample velocity/position algorithms derived from the associated integration formula. Simple comparison with existing velocity/position algorithms is performed in the context of level-flight examples in Section IV.

[1] Somewhat obscurely presented in the classic literature [1, 4, 5], it is difficult to fully understand for non-professionals and even for professionals.

Conclusions are drawn in Section V.

## II. Incremental Velocity/Position Integration Formulae

Denote by $N$ the local level navigation frame, by $B$ the body frame of the inertial navigation system, by $I$ the inertially non-rotating frame, by $E$ the Earth frame. The velocity and position rate equations in the navigation $N$-frame are respectively known as [1-3]

$$\dot{\mathbf{v}}^n = \mathbf{C}_b^n \mathbf{f}^b - \left(2\boldsymbol{\omega}_{ie}^n + \boldsymbol{\omega}_{en}^n\right) \times \mathbf{v}^n + \mathbf{g}^n \tag{1}$$

$$\dot{\mathbf{p}} = \mathbf{R}_c \mathbf{v}^n \tag{2}$$

where $\mathbf{v}^n$ denotes the vehicle's velocity relative to the Earth (also called ground velocity), $\mathbf{C}_b^n$ the attitude matrix from the body frame to the reference frame, $\mathbf{f}^b$ the specific force measured by accelerometers in $B$-frame, $\boldsymbol{\omega}_{ie}^n$ the Earth rotation rate with respect to the inertial frame, $\boldsymbol{\omega}_{en}^n$ the angular rate of the navigation frame with respect to the Earth frame, and $\mathbf{g}^n$ is the gravity vector. The vehicle's position $\mathbf{p} \triangleq \begin{bmatrix} \lambda & L & h \end{bmatrix}^T$ is described by the height above the Earth surface $h$ and the angular orientation of the navigation frame relative to the Earth frame, commonly expressed as longitude $\lambda$ and latitude $L$. $\mathbf{R}_c$ is the local curvature matrix that is a function of the current position. All the quantities above are functions of time and, if not explicitly stated, their time dependences on are omitted for brevity.

Next, we will consider the velocity and position updates from time $t_k$ to $t_{k+1}$ ($t_{k+1} - t_k \triangleq T$).

### A. Velocity Integration Formula

By the chain rule of the attitude matrix, $\mathbf{C}_b^n$ at any time

$$\mathbf{C}_{b(t)}^{n(t)} = \mathbf{C}_{n(t_k)}^{n(t)} \mathbf{C}_{b(t_k)}^{n(t_k)} \mathbf{C}_{b(t)}^{b(t_k)} \tag{3}$$

Both the body frame and the navigation frame with respect to any $I$-frame, say $\mathbf{C}_i^{b(t_k)}$ and $\mathbf{C}_i^{n(t_k)}$, are functions of $t_k$ instead of $t$, and hence their time derivatives are zero. To put it the other way, they are inertially "frozen" after the time epoch $t_k$ passes. Substituting (3) into (1) yields

$$\dot{\mathbf{v}}^n = \mathbf{C}_{n(t_k)}^{n(t)} \mathbf{C}_{b(t_k)}^{n(t_k)} \mathbf{C}_{b(t)}^{b(t_k)} \mathbf{f}^b - \left(2\boldsymbol{\omega}_{ie}^n + \boldsymbol{\omega}_{en}^n\right) \times \mathbf{v}^n + \mathbf{g}^n \tag{4}$$

Multiplying $\mathbf{C}_{n(t)}^{n(t_k)}$ on both sides,

$$\mathbf{C}_{n(t)}^{n(t_k)}\dot{\mathbf{v}}^n = \mathbf{C}_{b(t_k)}^{n(t_k)}\mathbf{C}_{b(t)}^{b(t_k)}\mathbf{f}^b - \mathbf{C}_{n(t)}^{n(t_k)}\left(2\boldsymbol{\omega}_{ie}^n + \boldsymbol{\omega}_{en}^n\right)\times\mathbf{v}^n + \mathbf{C}_{n(t)}^{n(t_k)}\mathbf{g}^n \tag{5}$$

Integrating over the time interval of interest,

$$\int_{t_k}^{t_{k+1}}\mathbf{C}_{n(t)}^{n(t_k)}\dot{\mathbf{v}}^n dt = \mathbf{C}_{b(t_k)}^{n(t_k)}\int_{t_k}^{t_{k+1}}\mathbf{C}_{b(t)}^{b(t_k)}\mathbf{f}^b dt - \int_{t_k}^{t_{k+1}}\mathbf{C}_{n(t)}^{n(t_k)}\left(2\boldsymbol{\omega}_{ie}^n + \boldsymbol{\omega}_{en}^n\right)\times\mathbf{v}^n dt + \int_{t_k}^{t_{k+1}}\mathbf{C}_{n(t)}^{n(t_k)}\mathbf{g}^n dt \tag{6}$$

The term on the left is derived as

$$\int_{t_k}^{t_{k+1}}\mathbf{C}_{n(t)}^{n(t_k)}\dot{\mathbf{v}}^n dt = \mathbf{C}_{n(t)}^{n(t_k)}\mathbf{v}^n\Big|_{t_k}^{t_{k+1}} - \int_{t_k}^{t_{k+1}}\mathbf{C}_{n(t)}^{n(t_k)}\boldsymbol{\omega}_{in}^n\times\mathbf{v}^n dt = \mathbf{C}_{n(t_{k+1})}^{n(t_k)}\mathbf{v}^n\left(t_{k+1}\right) - \mathbf{v}^n\left(t_k\right) - \int_{t_k}^{t_{k+1}}\mathbf{C}_{n(t)}^{n(t_k)}\boldsymbol{\omega}_{in}^n\times\mathbf{v}^n dt \tag{7}$$

where the attitude rate equation $\dot{\mathbf{C}}_n^i = \mathbf{C}_n^i\boldsymbol{\omega}_{in}^n\times$ is used. The skew symmetric matrix $(\cdot\times)$ is defined so that the cross product satisfies $\mathbf{p}\times\mathbf{q} = (\mathbf{p}\times)\mathbf{q}$ for arbitrary two vectors. Substituting (7) into (6) and reorganizing the terms,

$$\mathbf{v}^n\left(t_{k+1}\right) = \mathbf{C}_{n(t_k)}^{n(t_{k+1})}\left[\mathbf{v}^n\left(t_k\right) + \mathbf{C}_{b(t_k)}^{n(t_k)}\int_{t_k}^{t_{k+1}}\mathbf{C}_{b(t)}^{b(t_k)}\mathbf{f}^b dt - \int_{t_k}^{t_{k+1}}\mathbf{C}_{n(t)}^{n(t_k)}\boldsymbol{\omega}_{ie}^n\times\mathbf{v}^n dt + \int_{t_k}^{t_{k+1}}\mathbf{C}_{n(t)}^{n(t_k)}\mathbf{g}^n dt\right] \tag{8}$$

which is the analytic velocity integration formula in the potentially rotating navigation *N*-frame. Multiplication of the matrix $\mathbf{C}_{n(t_k)}^{n(t_{k+1})}$ depicts the rotating frame effect on the calculation of the velocity during the interval. The second term in the bracket $\mathbf{C}_{b(t_k)}^{n(t_k)}\int_{t_k}^{t}\mathbf{C}_{b(t)}^{b(t_k)}\mathbf{f}^b dt \triangleq \mathbf{u}(t)$ is the integration of the transformed specific force that necessitates the well-known sculling correction due to the body rotation [1, 5]. The last two terms introduce two new but similar integrals that can be handled by the sculling-like technique to account for the navigation frame rotation. In contrast to (8), previous works unexceptionally reckoned on some kinds of approximation, see e.g., Section 11.3-11.4 in [2], (10) in [5], and (7.2.2-1) and (7.2.2-1a) in [1], and (11.60). These approximations lead to their respective approximate algorithms. Details are provided in (21)-(23) in Section III-A.

### *B. Position Integration Formula*

In the context of a specific local level frame choice, e.g., North-Up-East, the local curvature matrix is explicitly expressed as a function of current position

$$\mathbf{R}_c = \begin{bmatrix} 0 & 0 & \dfrac{1}{(R_E + h)\cos L} \\ \dfrac{1}{R_N + h} & 0 & 0 \\ 0 & 1 & 0 \end{bmatrix} \tag{9}$$

where $R_E$ and $R_N$ are respectively the transverse radius of curvature and the meridian radius of curvature of the WGS-84 reference ellipsoid. The expression of $\mathbf{R}_c$ will be different for other local level frame choices, which, however, will not hinder from understanding the main idea of this paper. Clearly, $\mathbf{R}_c$ will encounter mathematical singularity when the cosine of the latitude approaches zero. In such rare cases, the angular orientation matrix of the navigation frame relative to the Earth frame can be used to encode the longitude and latitude information [1, 2]. It should be highlighted that the following development also applies after a little alternation.

As for the position update, integrating (2) from time $t_k$ to $t_{k+1}$ gives

$$\mathbf{p}(t_{k+1}) = \mathbf{p}(t_k) + \int_{t_k}^{t_{k+1}} \mathbf{R}_c \mathbf{v}^n dt \approx \mathbf{p}(t_k) + \mathbf{R}_c \int_{t_k}^{t_{k+1}} \mathbf{v}^n dt \tag{10}$$

where $\mathbf{R}_c$ is approximately taken as a constant and can be evaluated at, e.g., $t_k$, because the position changes very slowly during the integration interval. For $t \in [t_k, \quad t_{k+1}]$, define the position in *N*-frame as

$$\mathbf{r}^n(t) = \int_{t_k}^{t} \mathbf{v}^n dt \tag{11}$$

whose rate equation is given as

$$\dot{\mathbf{r}}^n = \mathbf{v}^n \tag{12}$$

Note that $\mathbf{r}^n(t_k) = 0$. The explicit form of $\mathbf{r}^n$ can be achieved by the similar technique as in deriving the velocity integration formula.

Substituting $t_{k+1}$ in (8) by $t$, (12) yields

$$\mathbf{C}_{n(t)}^{n(t_k)} \dot{\mathbf{r}}^n = \mathbf{C}_{n(t)}^{n(t_k)} \mathbf{v}^n = \mathbf{v}^n(t_k) + \mathbf{C}_{b(t_k)}^{n(t_k)} \int_{t_k}^{t} \mathbf{C}_{b(t)}^{b(t_k)} \mathbf{f}^b dt - \int_{t_k}^{t} \mathbf{C}_{n(t)}^{n(t_k)} \boldsymbol{\omega}_{ie}^n \times \mathbf{v}^n dt + \int_{t_k}^{t} \mathbf{C}_{n(t)}^{n(t_k)} \mathbf{g}^n dt \tag{13}$$

By the same techniques as in (7),

$$\int_{t_k}^{t_{k+1}} \mathbf{C}_{n(t)}^{n(t_k)} \dot{\mathbf{r}}^n dt = \mathbf{C}_{n(t)}^{n(t_k)} \mathbf{r}^n \Big|_{t_k}^{t_{k+1}} - \int_{t_k}^{t_{k+1}} \mathbf{C}_{n(t)}^{n(t_k)} \boldsymbol{\omega}_{in}^n \times \mathbf{r}^n dt = \mathbf{C}_{n(t_{k+1})}^{n(t_k)} \mathbf{r}(t_{k+1}) - \int_{t_k}^{t_{k+1}} \mathbf{C}_{n(t)}^{n(t_k)} \boldsymbol{\omega}_{in}^n \times \mathbf{r}^n dt \tag{14}$$

Integrating (13) over the time interval $\left[t_k,\quad t_{k+1}\right]$ and substituting (14), we obtain

$$\begin{aligned}\mathbf{r}\left(t_{k+1}\right)=\mathbf{C}_{n(t_k)}^{n(t_{k+1})}\Bigg[&T\mathbf{v}^n\left(t_k\right)+\int_{t_k}^{t_{k+1}}\mathbf{C}_{n(t)}^{n(t_k)}\boldsymbol{\omega}_{in}^n\times\mathbf{r}^n dt\\&+\mathbf{C}_{b(t_k)}^{n(t_k)}\int_{t_k}^{t_{k+1}}\int_{t_k}^{t}\mathbf{C}_{b(\tau)}^{b(t_k)}\mathbf{f}^b d\tau dt-\int_{t_k}^{t_{k+1}}\int_{t_k}^{t}\mathbf{C}_{n(\tau)}^{n(t_k)}\boldsymbol{\omega}_{ie}^n\times\mathbf{v}^n d\tau dt+\int_{t_k}^{t_{k+1}}\int_{t_k}^{t}\mathbf{C}_{n(\tau)}^{n(t_k)}\mathbf{g}^n d\tau dt\Bigg]\end{aligned} \tag{15}$$

which is the analytic position integration formula in the potentially rotating navigation *N*-frame. It consists of a single integral, the third term in the bracket, of the same structure with those in (8). The first term in the last row $\int_{t_k}^{t_{k+1}}\mathbf{u}\left(t\right)dt\triangleq I_{\mathbf{u}}\left(t_{k+1}\right)$ is the double integration of the transformed specific force in which the scrolling correction has to be applied to account for the body rotation in high-accuracy positioning applications [5]. By analogy, the last two double integrals can be handled by the scrolling-like correction to account for the navigation frame rotation. So far, the position algorithms in the literature all have employed some approximation forms, see e.g. the high-resolution position computation in (76)-(79) in [5].

The incremental velocity integration formula (8) and the incremental position integration formula (15) are settled as well-founded analytic cornerstones for the navigation computation algorithm design, and different approximations to the integrals involved will give birth to various velocity and position update algorithms. The velocity/position integration formulae cast the navigation computation algorithm design within a systematic and rather straightforward framework. In contrast to the previous works [1-3, 5], the navigation frame rotation effects are rigorously considered through the two integration formulae.

## III. Derived Velocity/Position Update Algorithms

This section uses the incremental velocity/position integration formulae as a basis to exemplify the design of the velocity/position update algorithms. Without loss of generality, the update time interval $\left[t_k,\quad t_{k+1}\right]$ is assumed to contain two samples of the gyroscope and accelerometer triads.

### *A. Velocity Update Algorithm*

Since $\boldsymbol{\omega}_{in}^n$ is usually a slowly changing quantity, it is reasonable to approximate the attitude matrix by $\mathbf{C}_{n(t)}^{n(t_k)}\approx I+\boldsymbol{\varphi}_n\times$, where $\boldsymbol{\varphi}_n\approx\left(t-t_k\right)\boldsymbol{\omega}_{in}^n$ denotes the *N*-frame rotation vector from $t_k$ to the current time. The last two integrals in the velocity integration formula (8) are respectively approximated by

$$\int_{t_k}^{t_{k+1}} \mathbf{C}_{n(t)}^{n(t_k)} \boldsymbol{\omega}_{ie}^n \times \mathbf{v}^n dt \approx \int_{t_k}^{t_{k+1}} \left( I + \left( t - t_k \right) \boldsymbol{\omega}_{in}^n \times \right) \boldsymbol{\omega}_{ie}^n \times \mathbf{v}^n dt \approx \left( TI + \frac{T^2}{2} \boldsymbol{\omega}_{in}^n \times \right) \boldsymbol{\omega}_{ie}^n \times \mathbf{v}^n$$
$$\int_{t_k}^{t_{k+1}} \mathbf{C}_{n(t)}^{n(t_k)} \mathbf{g}^n dt \approx \int_{t_k}^{t_{k+1}} \left( I + \left( t - t_k \right) \boldsymbol{\omega}_{in}^n \times \right) \mathbf{g}^n dt \approx \left( TI + \frac{T^2}{2} \boldsymbol{\omega}_{in}^n \times \right) \mathbf{g}^n \tag{16}$$

where the quantities $\boldsymbol{\omega}_{in}^n$, $\boldsymbol{\omega}_{ie}^n$, $\mathbf{v}^n$ and $\mathbf{g}^n$ are approximately regarded as constants and evaluated at $t_k$. Using the two-sample sculling correction, $\mathbf{u}\left(t_{k+1}\right)$ is usually approximated by (See the companion paper [12], Appendix A)

$$\mathbf{u}\left(t_{k+1}\right) = \mathbf{C}_{b(t_k)}^{n(t_k)} \left( \Delta\mathbf{v}_1 + \Delta\mathbf{v}_2 + \frac{1}{2}\left(\Delta\boldsymbol{\theta}_1 + \Delta\boldsymbol{\theta}_2\right) \times \left(\Delta\mathbf{v}_1 + \Delta\mathbf{v}_2\right) + \frac{2}{3}\left(\Delta\boldsymbol{\theta}_1 \times \Delta\mathbf{v}_2 + \Delta\mathbf{v}_1 \times \Delta\boldsymbol{\theta}_2\right) \right) \tag{17}$$

where $\Delta\mathbf{v}_1, \Delta\mathbf{v}_2$ are the first and second samples of the accelerometer-measured incremental velocity and $\Delta\boldsymbol{\theta}_1, \Delta\boldsymbol{\theta}_2$ are the first and second samples of the gyroscope-measured incremental angle, respectively.

Substituting into the velocity integration formula (8),

$$\mathbf{v}^n\left(t_{k+1}\right) \approx \mathbf{C}_{n(t_k)}^{n(t_{k+1})} \left\{ \mathbf{v}^n\left(t_k\right) + \mathbf{u}\left(t_{k+1}\right) - \left( TI + \frac{T^2}{2} \boldsymbol{\omega}_{in}^n \times \right) \boldsymbol{\omega}_{ie}^n \times \mathbf{v}^n\left(t_k\right) + \left( TI + \frac{T^2}{2} \boldsymbol{\omega}_{in}^n \times \right) \mathbf{g}^n \right\} \tag{18}$$

With the obtained $\mathbf{v}^n\left(t_{k+1}\right)$, the second integral in (8) can be refined to give a better approximation. Suppose the velocity changes linearly during $\left[t_k, \quad t_{k+1}\right]$, i.e.,

$$\mathbf{v}^n\left(t\right) = \mathbf{v}^n\left(t_k\right) + \frac{t - t_k}{T}\left(\mathbf{v}^n\left(t_{k+1}\right) - \mathbf{v}^n\left(t_k\right)\right) \tag{19}$$

then the first equation in (16) can be better approximated by

$$\begin{aligned} \int_{t_k}^{t_{k+1}} \mathbf{C}_{n(t)}^{n(t_k)} \boldsymbol{\omega}_{ie}^n \times \mathbf{v}^n dt &\approx \int_{t_k}^{t_{k+1}} \left( I + \left( t - t_k \right) \boldsymbol{\omega}_{in}^n \times \right) \boldsymbol{\omega}_{ie}^n \times \left( \mathbf{v}^n\left(t_k\right) + \frac{t - t_k}{T}\left(\mathbf{v}^n\left(t_{k+1}\right) - \mathbf{v}^n\left(t_k\right)\right) \right) dt \\ &\approx \left( TI + \frac{T^2}{2} \boldsymbol{\omega}_{in}^n \times \right) \boldsymbol{\omega}_{ie}^n \times \mathbf{v}^n\left(t_k\right) + \left( \frac{T}{2} I + \frac{T^2}{3} \boldsymbol{\omega}_{in}^n \times \right) \boldsymbol{\omega}_{ie}^n \times \left(\mathbf{v}^n\left(t_{k+1}\right) - \mathbf{v}^n\left(t_k\right)\right) \\ &= \left( \frac{T}{2} I + \frac{T^2}{6} \boldsymbol{\omega}_{in}^n \times \right) \boldsymbol{\omega}_{ie}^n \times \mathbf{v}^n\left(t_k\right) + \left( \frac{T}{2} I + \frac{T^2}{3} \boldsymbol{\omega}_{in}^n \times \right) \boldsymbol{\omega}_{ie}^n \times \mathbf{v}^n\left(t_{k+1}\right) \end{aligned} \tag{20}$$

Typical approximate velocity algorithms in the literature are presented below for easy reference. Totally ignoring the navigation frame rotation, the velocity algorithm in [2] gave

$$\mathbf{v}_{TN}^n\left(t_{k+1}\right) = \mathbf{v}^n\left(t_k\right) + \mathbf{u}\left(t_{k+1}\right) - T\left(2\boldsymbol{\omega}_{ie}^n + \boldsymbol{\omega}_{en}^n\right) \times \mathbf{v}^n\left(t_k\right) + T\mathbf{g}^n \tag{21}$$

A coarse compensation was proposed in [5], by replacing (6) with (10) therein, as

$$\mathbf{v}_{SV1}^n\left(t_{k+1}\right)=\mathbf{v}^n\left(t_k\right)+\left(I-T\boldsymbol{\omega}_{in}^n\times\right)\mathbf{u}\left(t_{k+1}\right)-T\left(2\boldsymbol{\omega}_{ie}^n+\boldsymbol{\omega}_{en}^n\right)\times\mathbf{v}^n\left(t_k\right)+T\mathbf{g}^n \tag{22}$$

With the assumptions of constant changing $\mathbf{C}_{n(t_k)}^{n(t)}$ and linearly ramping $\mathbf{u}(t)$, an improved velocity algorithm in [1, 13] was given as

$$\mathbf{v}_{SV2}^n\left(t_{k+1}\right)=\mathbf{v}^n\left(t_k\right)+\frac{1}{2}\left(\mathbf{C}_{n(t_k)}^{n(t_{k+1})}+I\right)\mathbf{u}\left(t_{k+1}\right)-T\left(2\boldsymbol{\omega}_{ie}^n+\boldsymbol{\omega}_{en}^n\right)\times\mathbf{v}^n\left(t_k\right)+T\mathbf{g}^n \tag{23}$$

It can be proved, however, that the above assumptions are rarely satisfied in practice (see Appendix for details).

*B. Position Update Algorithm*

Following (16) and noting that $\mathbf{r}^n\left(t_k\right)=0$, the single integral in the position integration formula (15)

$$\int_{t_k}^{t_{k+1}}\mathbf{C}_{n(t)}^{n(t_k)}\boldsymbol{\omega}_{in}^n\times\mathbf{r}^n dt\approx 0 \tag{24}$$

The last double integral is approximated by

$$\begin{aligned}&\int_{t_k}^{t_{k+1}}\int_{t_k}^{t}\mathbf{C}_{n(\tau)}^{n(t_k)}\mathbf{g}^n d\tau dt\approx\int_{t_k}^{t_{k+1}}\int_{t_k}^{t}\left(I+\left(\tau-t_k\right)\boldsymbol{\omega}_{in}^n\times\right)\mathbf{g}^n d\tau dt\\&\approx\int_{t_k}^{t_{k+1}}\left(\left(t-t_k\right)I+\frac{\left(t-t_k\right)^2}{2}\boldsymbol{\omega}_{in}^n\times\right)\mathbf{g}^n dt\\&\approx\left(\frac{T^2}{2}I+\frac{T^3}{6}\boldsymbol{\omega}_{in}^n\times\right)\mathbf{g}^n\end{aligned} \tag{25}$$

Using the linear velocity assumption (19), the second double integral is calculated by

$$\begin{aligned}&\int_{t_k}^{t_{k+1}}\int_{t_k}^{t}\mathbf{C}_{n(\tau)}^{n(t_k)}\boldsymbol{\omega}_{ie}^n\times\mathbf{v}^n d\tau dt\approx\int_{t_k}^{t_{k+1}}\int_{t_k}^{t}\left(I+\left(\tau-t_k\right)\boldsymbol{\omega}_{in}^n\times\right)\boldsymbol{\omega}_{ie}^n\times\left(\mathbf{v}^n\left(t_k\right)+\frac{\tau-t_k}{T}\left(\mathbf{v}^n\left(t_{k+1}\right)-\mathbf{v}^n\left(t_k\right)\right)\right)d\tau dt\\&\approx\left(\frac{T^2}{2}I+\frac{T^3}{6}\boldsymbol{\omega}_{in}^n\times\right)\boldsymbol{\omega}_{ie}^n\times\mathbf{v}^n\left(t_k\right)+\left(\frac{T^2}{6}I+\frac{T^3}{12}\boldsymbol{\omega}_{in}^n\times\right)\boldsymbol{\omega}_{ie}^n\times\left(\mathbf{v}^n\left(t_{k+1}\right)-\mathbf{v}^n\left(t_k\right)\right)\\&=\left(\frac{T^2}{3}I+\frac{T^3}{12}\boldsymbol{\omega}_{in}^n\times\right)\boldsymbol{\omega}_{ie}^n\times\mathbf{v}^n\left(t_k\right)+\left(\frac{T^2}{6}I+\frac{T^3}{12}\boldsymbol{\omega}_{in}^n\times\right)\boldsymbol{\omega}_{ie}^n\times\mathbf{v}^n\left(t_{k+1}\right)\end{aligned} \tag{26}$$

With the two-sample sculling correction, $I_{\mathbf{u}}\left(t_{k+1}\right)$ can be approximated by (See the companion paper [12], Appendix B)

$$I_{\mathbf{u}}\left(t_{k+1}\right)=\frac{T}{30}\mathbf{C}_{b(t_k)}^{n(t_k)}\left(25\Delta\mathbf{v}_1+5\Delta\mathbf{v}_2+12\Delta\boldsymbol{\theta}_1\times\Delta\mathbf{v}_1+8\Delta\boldsymbol{\theta}_1\times\Delta\mathbf{v}_2+2\Delta\mathbf{v}_1\times\Delta\boldsymbol{\theta}_2+2\Delta\boldsymbol{\theta}_2\times\Delta\mathbf{v}_2\right) \tag{27}$$

Substituting (24)-(26) into the position integration formula (15) yields

$$\mathbf{r}^n\left(t_{k+1}\right) \approx \mathbf{C}_{n(t_k)}^{n(t_{k+1})}\Big[T\mathbf{v}^n\left(t_k\right)+I_{\mathbf{u}}\left(t_{k+1}\right) \\ -\left(\frac{T^2}{3}I+\frac{T^3}{12}\boldsymbol{\omega}_{in}^n\times\right)\boldsymbol{\omega}_{ie}^n\times\mathbf{v}^n\left(t_k\right)-\left(\frac{T^2}{6}I+\frac{T^3}{12}\boldsymbol{\omega}_{in}^n\times\right)\boldsymbol{\omega}_{ie}^n\times\mathbf{v}^n\left(t_{k+1}\right)+\left(\frac{T^2}{2}I+\frac{T^3}{6}\boldsymbol{\omega}_{in}^n\times\right)\mathbf{g}^n\Big] \tag{28}$$

Modeling $\mathbf{r}^n\left(t\right)$ as a linear function of time with the obtained $\mathbf{r}^n\left(t_{k+1}\right)$, the single integral (24) can be refined as

$$\int_{t_k}^{t_{k+1}}\mathbf{C}_{n(t)}^{n(t_k)}\boldsymbol{\omega}_{in}^n\times\mathbf{r}^n dt \approx \left(\frac{T}{2}I+\frac{T^2}{3}\boldsymbol{\omega}_{in}^n\times\right)\boldsymbol{\omega}_{in}^n\times\mathbf{r}^n\left(t_{k+1}\right) \tag{29}$$

Previous works mostly use the position update by, e.g., the trapezoidal integration [2, 3]

$$\mathbf{r}_{TN}^n\left(t_{k+1}\right)=\frac{T}{2}\left(\mathbf{v}^n\left(t_k\right)+\mathbf{v}^n\left(t_{k+1}\right)\right) \tag{30}$$

By assuming linearly ramping $\mathbf{C}_{n(t_k)}^{n(t)}-I$ and $\mathbf{u}\left(t\right)$, a high-resolution position algorithm was proposed in [5] as

$$\mathbf{r}_{SV1}^n\left(t_{k+1}\right)=T\mathbf{v}^n\left(t_k\right)+I_{\mathbf{u}}\left(t_{k+1}\right)+\frac{T^2}{2}\left(\mathbf{g}^n-\left(2\boldsymbol{\omega}_{ie}^n+\boldsymbol{\omega}_{en}^n\right)\times\mathbf{v}^n\left(t_k\right)\right)+\frac{T}{3}\left(\mathbf{C}_{n(t_k)}^{n(t_{k+1})}-I\right)\mathbf{u}\left(t_{k+1}\right) \tag{31}$$

which was later refined in [1] to

$$\mathbf{r}_{SV2}^n\left(t_{k+1}\right)=T\mathbf{v}^n\left(t_k\right)+I_{\mathbf{u}}\left(t_{k+1}\right)+\frac{T^2}{2}\left(\mathbf{g}^n-\left(2\boldsymbol{\omega}_{ie}^n+\boldsymbol{\omega}_{en}^n\right)\times\mathbf{v}^n\left(t_k\right)\right)+\frac{T}{6}\left(\mathbf{C}_{n(t_k)}^{n(t_{k+1})}-I\right)\mathbf{u}\left(t_{k+1}\right) \tag{32}$$

The derived velocity/position algorithms in this paper and others aforementioned are summarized in Tables I-II for clear comparison.

## IV. Level-Flight Examples

The body rotation-induced algorithm errors are generally overwhelmed over the navigation frame rotation-induced algorithm errors, so we adopt the simple level flight examples in this paper, so as to make the latter kind of errors as much pronounced as possible for better comparison.

Let us first consider an airplane carrying with an inertial navigation system that flies level at a constant speed to the east. To make the analysis tractable, the body frame is assumed to be aligned with the local level frame during the whole flight. For this special level-flight case, the gyroscope-measured body angular rate is $\boldsymbol{\omega}_{ib}^b=\boldsymbol{\omega}_{in}^n=\boldsymbol{\omega}_{ie}^n+\boldsymbol{\omega}_{en}^n$ and the accelerometer-measured specific force $\mathbf{f}^b=\left(2\boldsymbol{\omega}_{ie}^n+\boldsymbol{\omega}_{en}^n\right)\times\mathbf{v}^n-\mathbf{g}^n$ according to (1). Note that $\boldsymbol{\omega}_{ib}^b$ and $\mathbf{f}^b$ are constant here. Consequently, (17) and (27) are respectively specified as $\mathbf{u} = T\left(I + \frac{T}{2}\boldsymbol{\omega}_{ib}^b \times\right)\mathbf{f}^b$ and $I_\mathbf{u} = \frac{T^2}{6}\left(3I + T\boldsymbol{\omega}_{ib}^b \times\right)\mathbf{f}^b$. Substituting into the velocity algorithms in (18), (21)-(23) gives, respectively,

$$\begin{aligned}
\mathbf{v}^n(t_{k+1}) &= \mathbf{C}_{n(t_k)}^{n(t_{k+1})}\left\{\mathbf{v}^n(t_k) + \left(TI + \frac{T^2}{2}\boldsymbol{\omega}_{in}^n \times\right)\left(\left(2\boldsymbol{\omega}_{ie}^n + \boldsymbol{\omega}_{en}^n\right)\times\mathbf{v}^n - \mathbf{g}^n\right) - \left(TI + \frac{T^2}{2}\boldsymbol{\omega}_{in}^n \times\right)\boldsymbol{\omega}_{ie}^n \times \mathbf{v}^n(t_k) + \left(TI + \frac{T^2}{2}\boldsymbol{\omega}_{in}^n \times\right)\mathbf{g}^n\right\} \\
&= \mathbf{C}_{n(t_k)}^{n(t_{k+1})}\left(I + T\boldsymbol{\omega}_{in}^n \times + \frac{T^2}{2}\left(\boldsymbol{\omega}_{in}^n \times\right)^2\right)\mathbf{v}^n(t_k) \\
&\approx \left(I - T\boldsymbol{\omega}_{in}^n \times + \frac{T^2}{2}\left(\boldsymbol{\omega}_{in}^n \times\right)^2\right)\left(I + T\boldsymbol{\omega}_{in}^n \times + \frac{T^2}{2}\left(\boldsymbol{\omega}_{in}^n \times\right)^2\right)\mathbf{v}^n(t_k) \\
&\approx \mathbf{v}^n(t_k) + \frac{T^4}{4}\left(\boldsymbol{\omega}_{in}^n \times\right)^4 \mathbf{v}^n(t_k)
\end{aligned} \tag{33}$$

$$\begin{aligned}
\mathbf{v}_{TN}^n(t_{k+1}) &= \mathbf{v}^n(t_k) + \frac{T^2}{2}\boldsymbol{\omega}_{in}^n \times\left(\left(2\boldsymbol{\omega}_{ie}^n + \boldsymbol{\omega}_{en}^n\right)\times\mathbf{v}^n(t_k) - \mathbf{g}^n\right) \\
&\approx \mathbf{v}^n(t_k) - \frac{T^2}{2}\boldsymbol{\omega}_{in}^n \times \mathbf{g}^n
\end{aligned} \tag{34}$$

$$\begin{aligned}
\mathbf{v}_{SV1}^n(t_{k+1}) &= \mathbf{v}^n(t_k) - \frac{T^2}{2}\boldsymbol{\omega}_{in}^n \times\left(\left(2\boldsymbol{\omega}_{ie}^n + \boldsymbol{\omega}_{en}^n\right)\times\mathbf{v}^n(t_k) - \mathbf{g}^n\right) \\
&\approx \mathbf{v}^n(t_k) + \frac{T^2}{2}\boldsymbol{\omega}_{in}^n \times \mathbf{g}^n
\end{aligned} \tag{35}$$

$$\begin{aligned}
\mathbf{v}_{SV2}^n(t_{k+1}) &= \frac{1}{2}\left(\mathbf{C}_{n(t_k)}^{n(t_{k+1})} + I\right)\mathbf{u}(t_{k+1}) + \left(I - T\left(2\boldsymbol{\omega}_{ie}^n + \boldsymbol{\omega}_{en}^n\right)\times\right)\mathbf{v}^n(t_k) + T\mathbf{g}^n \\
&\approx \mathbf{v}^n(t_k) + \frac{T^4}{8}\left(\boldsymbol{\omega}_{in}^n \times\right)^3 \mathbf{v}^n(t_k)
\end{aligned} \tag{36}$$

where $\mathbf{C}_{n(t_k)}^{n(t_{k+1})}$ are approximated up to two orders of the integration interval. It shows that, as far as one velocity update is concerned for a realistic airplane velocity, the derived algorithm is the most accurate, followed by the algorithm *SV2*. The other two are the same but with different signs.

For this special case, the position algorithm (28) is specified as

$$\begin{aligned}
\mathbf{r}^n(t_{k+1}) &\approx \mathbf{C}_{n(t_k)}^{n(t_{k+1})}\Bigg[T\mathbf{v}^n(t_k) + \frac{T^2}{6}\left(3I + T\boldsymbol{\omega}_{in}^n \times\right)\left(\left(2\boldsymbol{\omega}_{ie}^n + \boldsymbol{\omega}_{en}^n\right)\times\mathbf{v}^n - \mathbf{g}^n\right) \\
&\quad - \left(\frac{T^2}{3}I + \frac{T^3}{12}\boldsymbol{\omega}_{in}^n \times\right)\boldsymbol{\omega}_{ie}^n \times \mathbf{v}^n(t_k) - \left(\frac{T^2}{6}I + \frac{T^3}{12}\boldsymbol{\omega}_{in}^n \times\right)\boldsymbol{\omega}_{ie}^n \times \mathbf{v}^n(t_{k+1}) + \left(\frac{T^2}{2}I + \frac{T^3}{6}\boldsymbol{\omega}_{in}^n \times\right)\mathbf{g}^n\Bigg] \\
&= \mathbf{C}_{n(t_k)}^{n(t_{k+1})}\left(I + \frac{T}{2}\boldsymbol{\omega}_{in}^n \times + \frac{T^2}{6}\left(\boldsymbol{\omega}_{in}^n \times\right)^2\right)T\mathbf{v}^n(t_k)
\end{aligned}$$

where $\mathbf{v}^n(t_{k+1})$ is replaced by $\mathbf{v}^n(t_k)$ for simpler comparison. When (29) is incorporated into (15), $\mathbf{r}^n(t_{k+1})$ is refined to

$$\begin{aligned}&\left(I+\mathbf{C}_{n(t_k)}^{n(t_{k+1})}\left(\frac{T}{2}I+\frac{T^2}{3}\boldsymbol{\omega}_{in}^n\times\right)\boldsymbol{\omega}_{in}^n\times\right)\mathbf{r}^n(t_{k+1})\\&\approx T\mathbf{v}^n(t_k)-\frac{T^3}{4}\left(\boldsymbol{\omega}_{in}^n\times\right)^2\mathbf{v}^n(t_k)\end{aligned} \tag{37}$$

The other position algorithms (30)-(32) are respectively specified as

$$\mathbf{r}_{TN}^n(t_{k+1})=T\mathbf{v}^n(t_k) \tag{38}$$

$$\begin{aligned}\mathbf{r}_{SV1}^n(t_{k+1})&=T\mathbf{v}^n(t_k)+I_{\mathbf{u}}(t_{k+1})+\frac{T^2}{2}\left(\mathbf{g}^n-\left(2\boldsymbol{\omega}_{ie}^n+\boldsymbol{\omega}_{en}^n\right)\times\mathbf{v}^n(t_k)\right)+\frac{T}{3}\left(\mathbf{C}_{n(k)}^{n(k+1)}-I\right)\mathbf{u}(t_{k+1})\\&=T\mathbf{v}^n(t_k)-\frac{T^3}{6}\boldsymbol{\omega}_{in}^n\times\left(\left(2\boldsymbol{\omega}_{ie}^n+\boldsymbol{\omega}_{en}^n\right)\times\mathbf{v}^n(t_k)-\mathbf{g}^n\right)\\&\approx T\mathbf{v}^n(t_k)+\frac{T^3}{6}\boldsymbol{\omega}_{in}^n\times\mathbf{g}^n\end{aligned} \tag{39}$$

$$\begin{aligned}\mathbf{r}_{SV2}^n(t_{k+1})&=T\mathbf{v}^n(t_k)+I_{\mathbf{u}}(t_{k+1})+\frac{T^2}{2}\left(\mathbf{g}^n-\left(2\boldsymbol{\omega}_{ie}^n+\boldsymbol{\omega}_{en}^n\right)\times\mathbf{v}^n(t_k)\right)+\frac{T}{6}\left(\mathbf{C}_{n(t_k)}^{n(t_{k+1})}-I\right)\mathbf{u}(t_{k+1})\\&=T\mathbf{v}^n(t_k)+\frac{T^5}{24}\left(\boldsymbol{\omega}_{in}^n\times\right)^3\left(\left(2\boldsymbol{\omega}_{ie}^n+\boldsymbol{\omega}_{en}^n\right)\times\mathbf{v}^n(t_k)-\mathbf{g}^n\right)\\&\approx T\mathbf{v}^n(t_k)-\frac{T^5}{24}\left(\boldsymbol{\omega}_{in}^n\times\right)^3\mathbf{g}^n\end{aligned} \tag{40}$$

As far as one position update is concerned in the considered example, the algorithm *TN* runs as the most accurate one, followed by *SV2*, the derived one and *SV1* in the accuracy-descending order. The specific velocity and position algorithms for the level-flight const-speed example are listed in the right columns of Tables I-II.

The above level-flight example is simulated for an hour with an east velocity 500 *m/s* at latitude $30^{\circ}$. The algorithm update interval is set to $T=0.02s$, with no vertical damping applied. The horizontal velocity errors and horizontal position errors for each algorithm are plotted in Figs. 1-2, respectively. The algorithms *TN* and *SV1* come with the same largest error behaviors, with the maximum position error of over ten meters. The derived algorithm shows the smallest error in both velocity and position, tightly followed by *SV2*. It should be emphasized that an infinitely small number $10^{-20}$ is used in Fig. 2 to represent the actual zero in the simulation result and the sudden jump of *SV2* at about 2800s is owed to the numerical truncation error. In this case, $\left|\boldsymbol{\omega}_{in}^n\right|=1.6\times10^{-4}$ and $\left|\boldsymbol{\omega}_{ib}^{b}\times\mathbf{f}^{b}+\dot{\mathbf{f}}^{b}\right|\approx\left|\boldsymbol{\omega}_{ib}^{b}\times\mathbf{f}^{b}\right|=0.0014$. The assumptions in *SV2* (see (41)-(42)) are roughly satisfied.

We further simulated and examined another level flight example with the east velocity rate $\dot{\mathbf{v}}_{E}^{n}=a\sin\left(wt\right)$. When the rate magnitude $a=10$ and the rate angular frequency $w=0.02\pi$, the east velocity profile at 0-300s is plotted in Fig. 3. The horizontal velocity and position errors (2-hour) for each algorithm are respectively given in Figs. 4-5. The derived algorithm is significantly the smallest in both velocity and position errors, followed by *SV2, TN* and *SV1* in the accuracy-descending order. We have also performed many other simulations by changing the velocity rate's magnitude and angular frequency, and the rank result is quite similar. For such an example, it can be shown that $\max\left(\left|\boldsymbol{\omega}_{in}^{n}\right|\right)=2.2\times10^{-4}$ and $\max\left(\left|\boldsymbol{\omega}_{ib}^{b}\times\mathbf{f}^{b}+\dot{\mathbf{f}}^{b}\right|\right)\approx\max\left(\left|\dot{\mathbf{f}}^{b}\right|\right)=0.63$, the latter of which badly violates the second assumption (42). It explains the unsatisfying behavior of *SV2* in Figs. 4-5.

## V. Conclusions

Navigation frame rotation is an important issue that should be well-considered in the future ultra-precision inertial navigation algorithm design, but has been less seriously handled so far. In this paper, the velocity and position integration formulae are employed to rigorously tackle the navigation frame rotation issue. In doing so, the inertial navigation velocity/position algorithms design is cast into a systematic and straightforward framework that hopefully benefits the comprehension of the inertial navigation computation principle. Different approximations to the integrals involved in the velocity/position integration formulae give birth to various velocity/position update algorithms. Two-sample velocity and position algorithms are derived to demonstrate the design process within the framework. In the context of level-flight airplane examples, the derived algorithm is analytically and numerically compared to the typical navigation algorithms in the literature. Significant benefits of the derived algorithms are observed.

## Appendix

Here we dwell upon the assumptions in deriving the velocity algorithm *SV2* in [1].

Since $\dot{\mathbf{C}}_{n(t_k)}^{n(t)}=-\left(\boldsymbol{\omega}_{in}^{n}\times\right)\mathbf{C}_{n(t_k)}^{n(t)}$ and rigid rotations do not change the length of a vector, "*constant changing* $\mathbf{C}_{n(t_k)}^{n(t)}$" means

$$0=\ddot{\mathbf{C}}_{n(t_k)}^{n(t)}=\left(\boldsymbol{\omega}_{in}^{n}\times\right)^2\mathbf{C}_{n(t_k)}^{n(t)}-\left(\dot{\boldsymbol{\omega}}_{in}^{n}\times\right)\mathbf{C}_{n(t_k)}^{n(t)}$$
$$\Leftrightarrow\left(\boldsymbol{\omega}_{in}^{n}\times\right)^2-\left(\dot{\boldsymbol{\omega}}_{in}^{n}\times\right)=0 \tag{41}$$

which is satisfied when $\boldsymbol{\omega}_{in}^{n}=0$, namely, *N*-frame is an inertial frame.

Similarly, since $\dot{\mathbf{C}}_{b(t)}^{b(t_k)}=\mathbf{C}_{b(t)}^{b(t_k)}\boldsymbol{\omega}_{ib}^{b}\times$ and $\dot{\mathbf{u}}(t)=\mathbf{C}_{b(t_k)}^{n(t_k)}\mathbf{C}_{b(t)}^{b(t_k)}\mathbf{f}^b$ (see the text below (8) for $\mathbf{u}(t)$'s definition), *"linearly ramping $\mathbf{u}(t)$"* means

$$0=\ddot{\mathbf{u}}(t)=\mathbf{C}_{b(t_k)}^{n(t_k)}\mathbf{C}_{b(t)}^{b(t_k)}\boldsymbol{\omega}_{ib}^{b}\times\mathbf{f}^b+\mathbf{C}_{b(t_k)}^{n(t_k)}\mathbf{C}_{b(t)}^{b(t_k)}\dot{\mathbf{f}}^b$$
$$\Leftrightarrow\boldsymbol{\omega}_{ib}^{b}\times\mathbf{f}^b+\dot{\mathbf{f}}^b=0 \tag{42}$$

which is valid only under rare conditions, for example, when the INS is rotated with zero origin translation (see (21) in [14]).

## ACKNOWLEDGEMENTS

Special thanks to the audiences of the university Graduate Course "Autonomous Navigation and Its Applications" in Spring 2011, who inspire the authors to consider the subject of this paper in a more serious manner.

## REFERENCES

[1] P. G. Savage, *Strapdown Analytics*, 2nd ed.: Strapdown Analysis, 2007.

[2] D. H. Titterton and J. L. Weston, *Strapdown Inertial Navigation Technology*: the Institute of Electrical Engineers, London, United Kingdom, 2nd Ed., 2004.

[3] P. D. Groves, *Principles of GNSS, Inertial, and Multisensor Integrated Navigation Systems*: Artech House, Boston and London, 2008.

[4] P. G. Savage, "Strapdown inertial navigation integration algorithm design, part 1: attitude algorithms," *Journal of Guidance, Control, and Dynamics,* vol. 21, pp. 19-28, 1998.

[5] P. G. Savage, "Strapdown inertial navigation integration algorithm design, part 2: velocity and position algorithms," *Journal of Guidance, Control, and Dynamics,* vol. 21, pp. 208-221, 1998.

[6] D. A. Tazarts, "Inertial Navigation: From Gimbaled Platforms to Strapdown Sensors," *IEEE Trans. on Aerospace and Electronic Systems,* vol. 47, pp. 2292-2299, 2011.

[7] Y. Wu*, et al.*, "Strapdown inertial navigation system algorithms based on dual quaternions," *IEEE Transactions on*

*Aerospace and Electronic Systems,* vol. 41, pp. 110-132, 2005.

[8] P. G. Savage, "A unified mathematical framework for strapdown algorithm design," *Journal of Guidance Control and Dynamics,* vol. 29, pp. 237-249, 2006.

[9] Y. Wu, "On "A unified mathematical framework for strapdown algorithm design"," *Journal of Guidance, Control, and Dynamics,* vol. 29, pp. 1482-1484, 2006.

[10] N. M. Barbour, "Inertial Navigation Sensors," in *Low-Cost Navigation Sensors and Integration Technology* ed: NATO RTO-EN-SET-116, http://www.rta.nato.int/pubs/rdp.asp?rdp=RTO-EN-SET-116, 2009.

[11] P. G. Savage, "Strapdown system algorithms," AGARD Lecture Series No. 133, 1984.

[12] Y. Wu and X. Pan, "Velocity/Position Integration Formula (I): Application to In-flight Alignment," *submitted to IEEE Trans. on Aerospace and Electronic Systems,* 2011.

[13] P. G. Savage, "Errata to "Strapdown inertial navigation integration algorithm design, part 2: velocity and position algorithms"," *Journal of Guidance, Control, and Dynamics,* vol. 27, p. 318, 2004.

[14] Y. Wu*, et al.*, "Observability of SINS Alignment: A Global Perspective," *IEEE Trans. on Aerospace and Electronic Systems,* vol. 48, pp. 78-102, 2012.

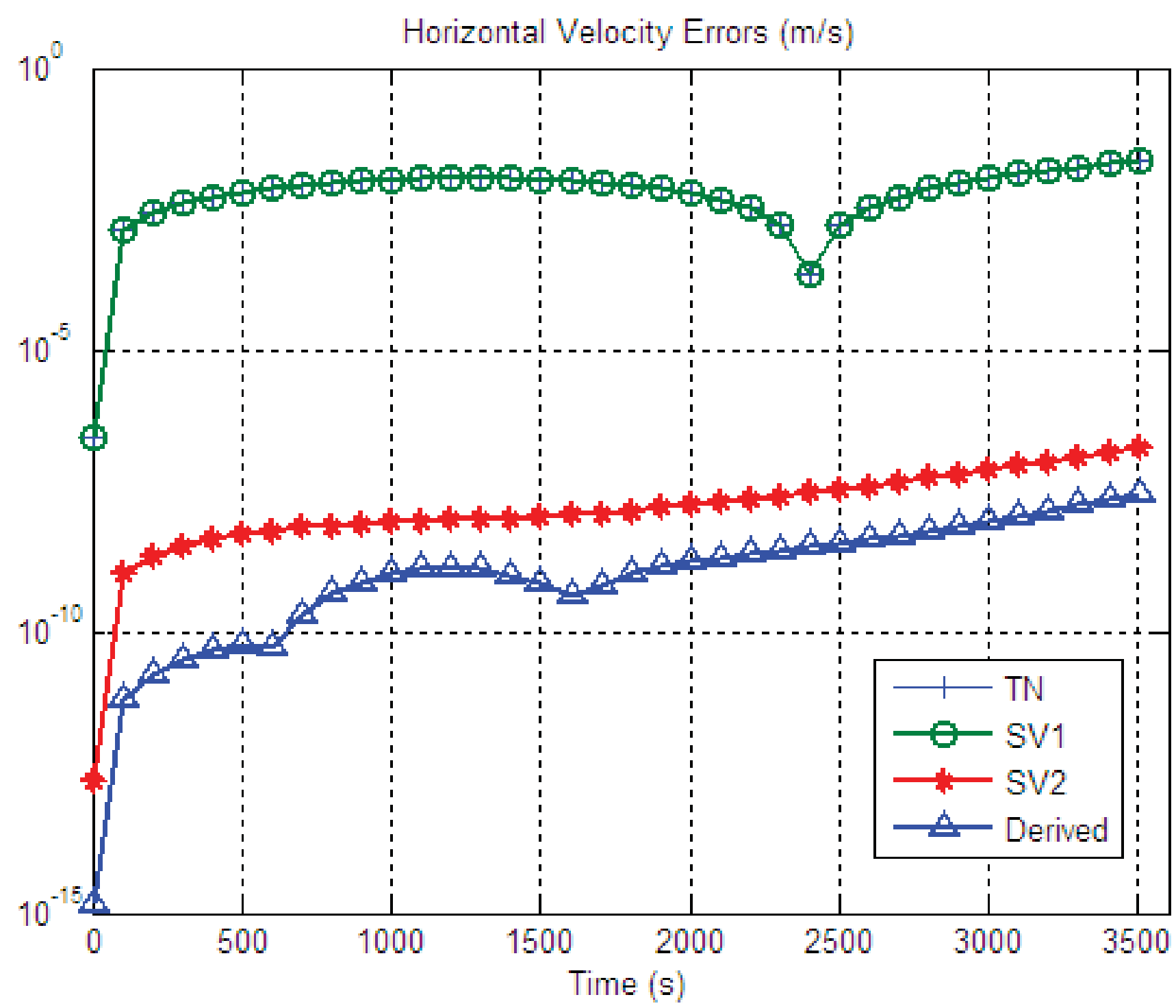

Figure 1. Horizontal velocity error comparison for level-flight const-velocity case

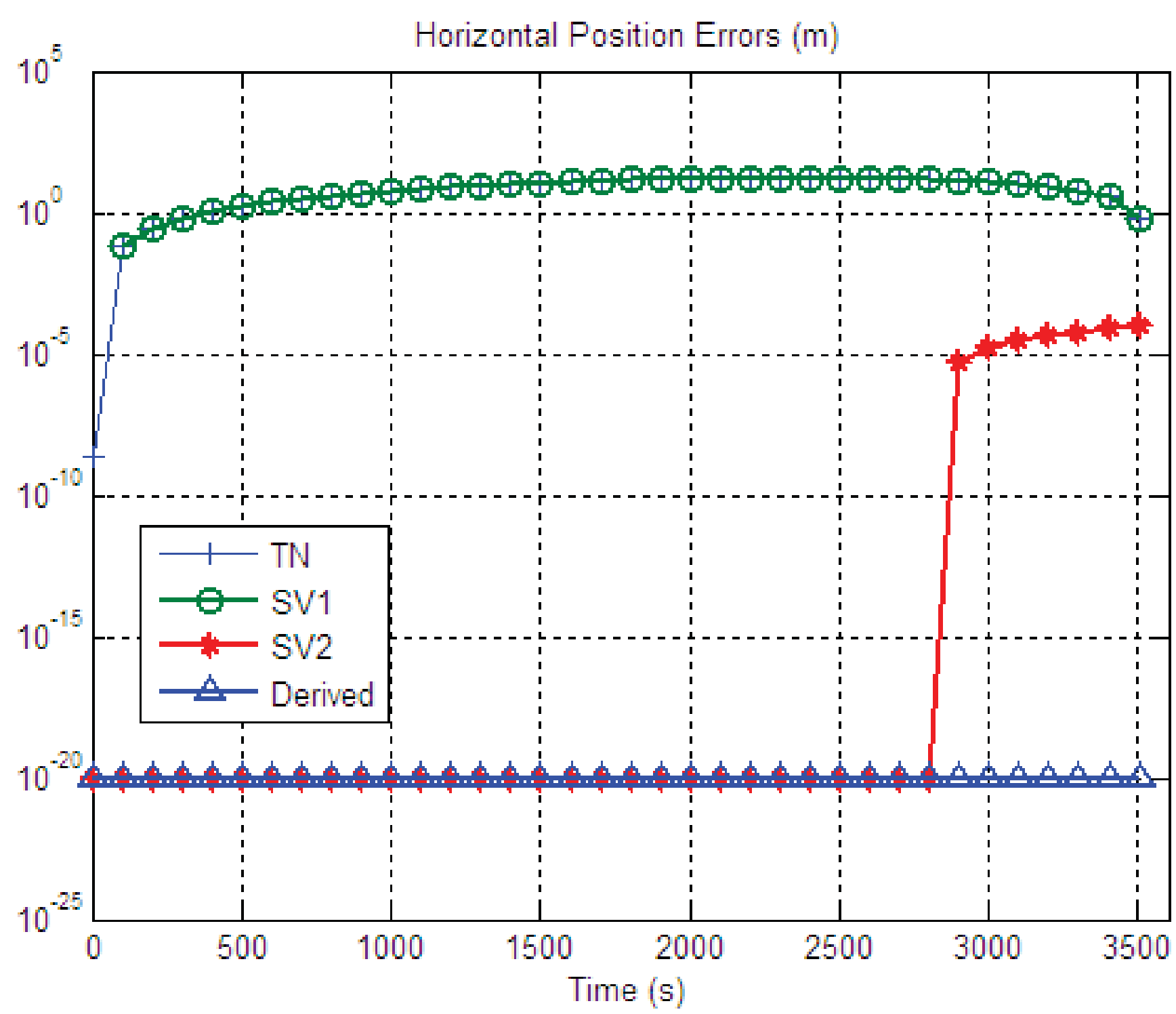

Figure 2. Horizontal position error comparison for level-flight const-velocity case

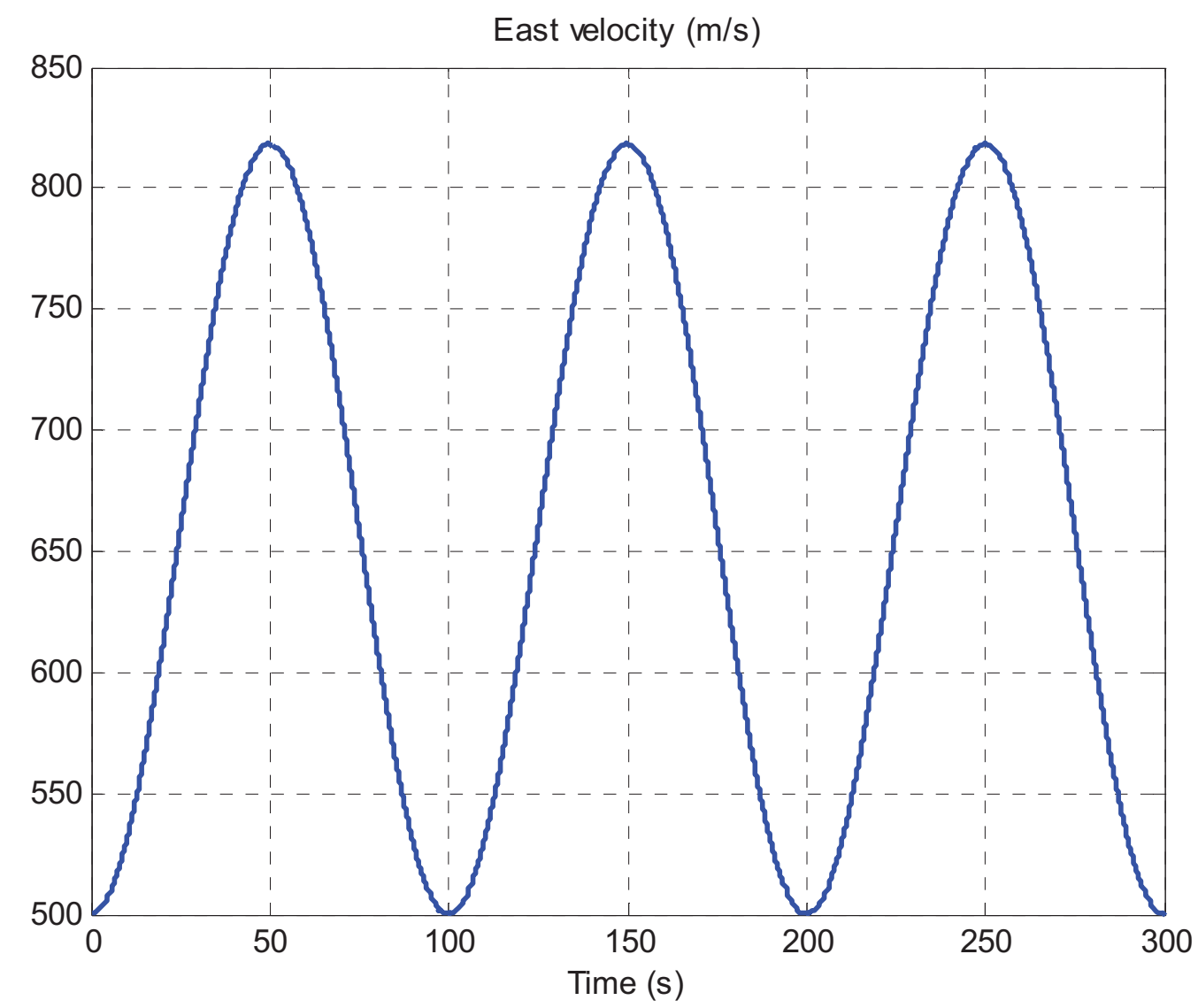

Figure 3. East velocity profile.

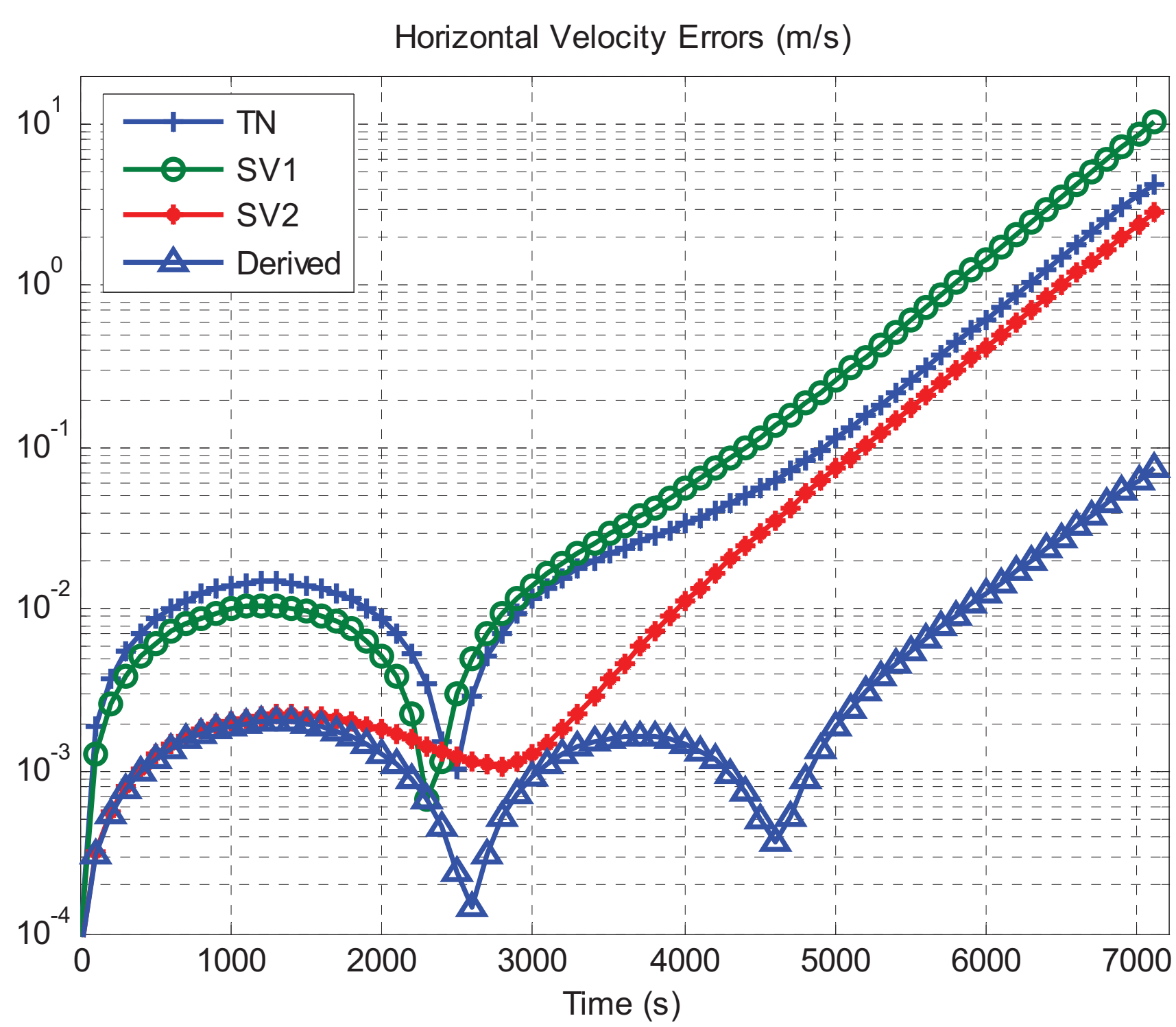

Figure 4. Horizontal velocity error comparison for level-flight varying-velocity case

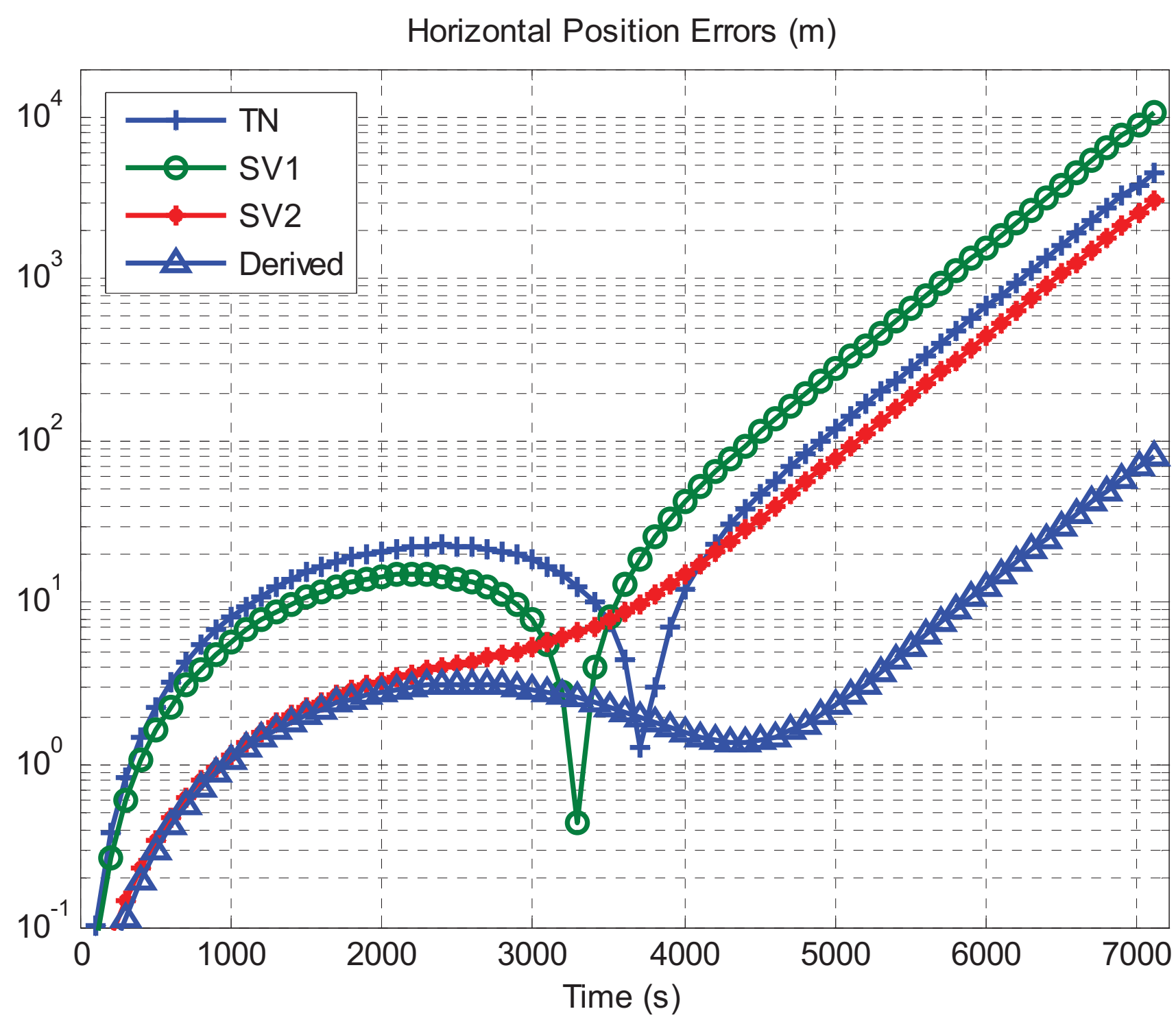

Figure 5. Horizontal position error comparison for level-flight varying -velocity case

Table I. Summary of Velocity Algorithms

| Velocity Algorithm | General Case | Level-flight Const-speed Case |
|---|---|---|
| ***The Derived*** | (18): $\mathbf{C}_{n(t_k)}^{n(t_{k+1})}\left\{\mathbf{v}^n(t_k)+\mathbf{u}(t_{k+1})-\left(TI+\frac{T^2}{2}\boldsymbol{\omega}_{in}^n\times\right)\boldsymbol{\omega}_{ie}^n\times\mathbf{v}^n(t_k)+\left(TI+\frac{T^2}{2}\boldsymbol{\omega}_{in}^n\times\right)\mathbf{g}^n\right\}$ | (33): $\mathbf{v}^n(t_k)+\frac{T^4}{4}\left(\boldsymbol{\omega}_{in}^n\times\right)^4\mathbf{v}^n(t_k)$ |
| ***TN*** | (21): $\mathbf{v}^n(t_k)+\mathbf{u}(t_{k+1})-T\left(2\boldsymbol{\omega}_{ie}^n+\boldsymbol{\omega}_{en}^n\right)\times\mathbf{v}^n(t_k)+T\mathbf{g}^n$ | (34): $\mathbf{v}^n(t_k)-\frac{T^2}{2}\boldsymbol{\omega}_{in}^n\times\mathbf{g}^n$ |
| ***SV1*** | (22): $\mathbf{v}^n(t_k)+\left(I-T\boldsymbol{\omega}_{in}^n\times\right)\mathbf{u}(t_{k+1})-T\left(2\boldsymbol{\omega}_{ie}^n+\boldsymbol{\omega}_{en}^n\right)\times\mathbf{v}^n(t_k)+T\mathbf{g}^n$ | (35): $\mathbf{v}^n(t_k)+\frac{T^2}{2}\boldsymbol{\omega}_{in}^n\times\mathbf{g}^n$ |
| ***SV2*** | (23): $\mathbf{v}^n(t_k)+\frac{1}{2}\left(\mathbf{C}_{n(t_k)}^{n(t_{k+1})}+I\right)\mathbf{u}(t_{k+1})-T\left(2\boldsymbol{\omega}_{ie}^n+\boldsymbol{\omega}_{en}^n\right)\times\mathbf{v}^n(t_k)+T\mathbf{g}^n$ | (36): $\mathbf{v}^n(t_k)+\frac{T^4}{8}\left(\boldsymbol{\omega}_{in}^n\times\right)^3\mathbf{v}^n(t_k)$ |

Table II. Summary of Position Algorithms

| Position Algorithm | General Case | Level-flight Const-speed Case |
|---|---|---|
| ***The Derived*** | (28): $\mathbf{C}_{n(t_k)}^{n(t_{k+1})}\Big[T\mathbf{v}^n(t_k)+I_{\mathbf{u}}(t_{k+1})$ $-\left(\frac{T^2}{3}I+\frac{T^3}{12}\boldsymbol{\omega}_{in}^n\times\right)\boldsymbol{\omega}_{ie}^n\times\mathbf{v}^n(t_k)-\left(\frac{T^2}{6}I+\frac{T^3}{12}\boldsymbol{\omega}_{in}^n\times\right)\boldsymbol{\omega}_{ie}^n\times\mathbf{v}^n(t_{k+1})+\left(\frac{T^2}{2}I+\frac{T^3}{6}\boldsymbol{\omega}_{in}^n\times\right)\mathbf{g}^n\Big]$ | (37): $T\mathbf{v}^n(t_k)-\frac{T^3}{4}\left(\boldsymbol{\omega}_{in}^n\times\right)^2\mathbf{v}^n(t_k)$ |
| ***TN*** | (30): $\frac{T}{2}\left(\mathbf{v}^n(t_k)+\mathbf{v}^n(t_{k+1})\right)$ | (38): $T\mathbf{v}^n(t_k)$ |
| ***SV1*** | (31): $T\mathbf{v}^n(t_k)+I_{\mathbf{u}}(t_{k+1})+\frac{T^2}{2}\left(\mathbf{g}^n-\left(2\boldsymbol{\omega}_{ie}^n+\boldsymbol{\omega}_{en}^n\right)\times\mathbf{v}^n(t_k)\right)+\frac{T}{3}\left(\mathbf{C}_{n(t_k)}^{n(t_{k+1})}-I\right)\mathbf{u}(t_{k+1})$ | (39): $T\mathbf{v}^n(t_k)+\frac{T^3}{6}\boldsymbol{\omega}_{in}^n\times\mathbf{g}^n$ |
| ***SV2*** | (32): $T\mathbf{v}^n(t_k)+I_{\mathbf{u}}(t_{k+1})+\frac{T^2}{2}\left(\mathbf{g}^n-\left(2\boldsymbol{\omega}_{ie}^n+\boldsymbol{\omega}_{en}^n\right)\times\mathbf{v}^n(t_k)\right)+\frac{T}{6}\left(\mathbf{C}_{n(t_k)}^{n(t_{k+1})}-I\right)\mathbf{u}(t_{k+1})$ | (40): $T\mathbf{v}^n(t_k)-\frac{T^5}{24}\left(\boldsymbol{\omega}_{in}^n\times\right)^3\mathbf{g}^n$ |